\documentclass[acmtog, authorversion,nonacm]{acmart}

\usepackage{booktabs} 

\usepackage[ruled]{algorithm2e} 

\SetAlFnt{\small}
\SetAlCapFnt{\small}
\SetAlCapNameFnt{\small}
\SetAlCapHSkip{0pt}

\begin{document}
\title{Assembler: Scalable 3D Part Assembly via Anchor Point Diffusion}

\author{Wang Zhao}
\affiliation{%
  \institution{Tencent ARC Lab}
  \country{China}}
\email{wangzhao@tencent.com}

\author{Yan-Pei Cao}
\affiliation{%
  \institution{VAST}
  \country{China}}
\email{caoyanpei@gmail.com}

\author{Jiale Xu}
\affiliation{%
  \institution{Tencent ARC Lab}
  \country{China}}
\email{jialexu@tencent.com}

\author{Yuejiang Dong}
\affiliation{%
  \institution{Tsinghua University}
  \country{China}}
\email{dongyj21@mails.tsinghua.edu.cn}

\author{Ying Shan}
\affiliation{%
  \institution{Tencent ARC Lab}
  \country{China}}
\email{yingsshan@tencent.com}

\begin{abstract}
We present Assembler, a scalable and generalizable framework for 3D part assembly that reconstructs complete objects from input part meshes and a reference image. Unlike prior approaches that mostly rely on deterministic part pose prediction and category-specific training, Assembler is designed to handle diverse, in-the-wild objects with varying part counts, geometries, and structures. It addresses the core challenges of scaling to general 3D part assembly through innovations in task formulation, representation, and data. First, Assembler casts part assembly as a generative problem and employs diffusion models to sample plausible configurations, effectively capturing ambiguities arising from symmetry, repeated parts, and multiple valid assemblies. Second, we introduce a novel shape-centric representation based on sparse anchor point clouds, enabling scalable generation in Euclidean space rather than SE(3) pose prediction. Third, we construct a large-scale dataset of over 320K diverse part-object assemblies using a synthesis and filtering pipeline built on existing 3D shape repositories. Assembler achieves state-of-the-art performance on PartNet and is the first to demonstrate high-quality assembly for complex, real-world objects. Based on Assembler, we further introduce an interesting part-aware 3D modeling system that generates high-resolution, editable objects from images, demonstrating potential for interactive and compositional design. \it{Project page: \url{https://assembler3d.github.io}}
\end{abstract}

%
%

%
%


\begin{teaserfigure}
    \centering
    \includegraphics[width=1\textwidth]{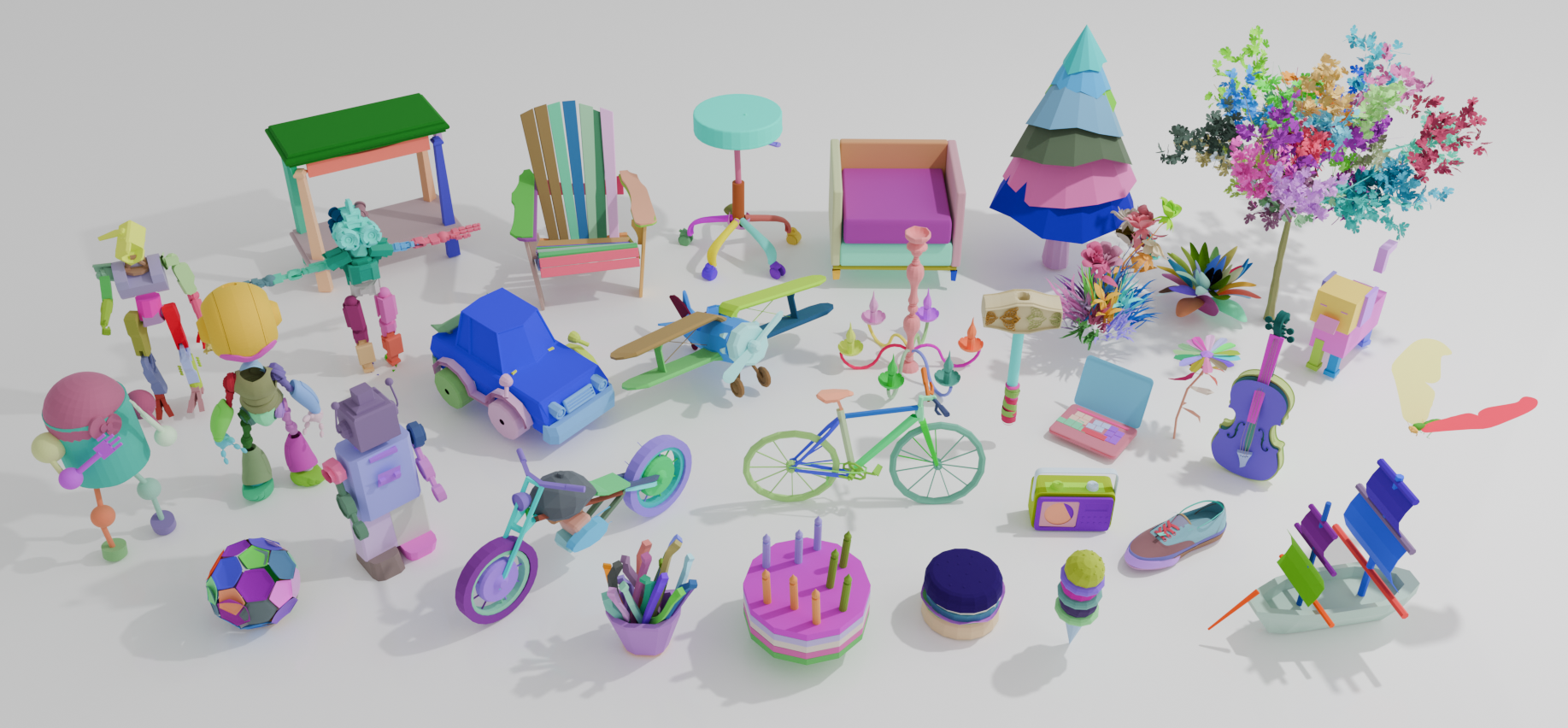}
    \caption{3D part assemblies of general objects by Assembler. Parts are labeled in different colors.}
    \label{fig:teaser}
\end{teaserfigure}

\maketitle

\section{Introduction}
3D part assembly is a fundamental task in computer vision and graphics, aiming at composing a complete object from a set of modular parts. This capability is increasingly critical across a wide range of applications, including 3D content creation, computer-aided design (CAD), manufacturing, and robotics. A robust, generalizable assembly system could greatly enhance 3D modeling workflows and unlock new possibilities for interactive and automated design.

Despite its importance, automatic 3D part assembly remains a highly challenging problem. It requires a comprehensive understanding of part geometry and semantics, the ability to reason about inter-part relationships, and the capacity to imagine plausible complete object shapes. Some works~\cite{huang2006reassembling, sellan2022breaking, lu2023jigsaw, xu2025high} have explored the 3D fracture assembly problem, which focus primarily on low-level geometric cues—e.g., boundary curves and local correspondences—to reassemble broken objects. In contrast, 3D part assembly typically assumes that each part is a complete, semantically meaningful unit and often permits duplicated components. These characteristics introduce unique challenges, demanding higher-level reasoning about object structure, part functionality, and overall semantic coherence.

Over the past decades, researchers have explored a variety of approaches to tackle the 3D part assembly problem. Early works addressed part assembly by developing efficient retrieval and registration based on curated part libraries~\cite{funkhouser2004modeling}, or constructing graphical models~\cite{chaudhuri2011probabilistic, kalogerakis2012probabilistic} to capture part semantics and relational constraints. While significantly reducing the manual efforts, they often relied on pre-defined part taxonomies or databases, limiting their scalability and generalization to novel or unstructured settings. With the rise of deep learning, recent works~\cite{zhan2020generative} have proposed to encode part features using neural networks, learning to predict the 6-DoF poses of parts to assemble. Following this direction, subsequent methods have introduced more powerful architectures~\cite{zhang20223d, zhang2024scalable, xu2025spaformer}, incorporated richer input information~\cite{narayan2022rgl, li2020learning}, and adopted hierarchical structures~\cite{du2024generative} to improve assembly accuracy. Although these models have achieved promising results—particularly on canonical object categories such as chairs, tables, and lamps from the PartNet dataset~\cite{mo2019partnet}—they remain largely constrained to category-specific scenarios with relatively simple, well-aligned parts. How to scale these methods to handle general, in-the-wild objects with diverse geometries, part counts, and structural variations remains an open challenge in the field.

In this work, we present Assembler, a scalable and effective framework for 3D part assembly of general objects. Given a set of part meshes and a reference image, Assembler produces accurate, high-fidelity assemblies that generalize across varying numbers of parts, intricate shapes, and diverse object categories. It comprehensively addresses the above scalability challenges of general 3D part assembly through three key innovations. 

First, we formulate 3D part assembly as a generative task, and leverage diffusion models to learn the distribution of plausible assemblies. A specially-designed diffusion model is introduced to effectively extract rich, aligned features from input conditions. By modeling as a generation task rather than the deterministic prediction, Assembler naturally handles the inherent ambiguities of part assembly—such as symmetry, duplicate components, and multiple valid configurations—thus better supporting generalization to in-the-wild objects. This shares similar motivations with recent works, which explored diffusion models for 2D~\cite{scarpellini2024diffassemble} and fracture~\cite{sellan2022breaking} assembly, and a score-based generative model for category-specific part assembly~\cite{cheng2023score}. 

Second, we propose a novel representation for part assemblies based on sparse anchor point clouds. Existing methods typically predict the 6-DoF poses of parts in SE(3). However, this parametric representation is highly abstract and geometric that lacks explicit assembled shape information. Furthermore, the probabilistic distribution of poses is difficult for generative modeling due to its discontinuities and multimodal distribution, as observed in ~\cite{leach2022denoising, yao2025cast}. In constrast, in Assembler, we innovatively introduce sparse anchor point clouds as the assembly representation, and formulate the part assembly as a point cloud generation problem. Each input part is sampled as a set of sparse anchor points, and the model generates an assembled anchor point cloud representing the final object shape. Part poses can be easily recovered by simple least-squares fitting. This shape-centric formulation enables the diffusion model to learn smooth and scalable distributions of the assembled object point clouds in Euclidean space, whose effectiveness is well validated in prior 3D point cloud generation methods~\cite{nichol2022point, lan2024ga}. 

Third, we address the lack of large-scale data for general 3D part assembly by proposing a data synthesis and filtering pipeline. Leveraging the disconnected-component property of existing large-scale 3D shape datasets, we generate diverse part-object pairs through segmentation and filtering. We collect and construct a dataset of over 320,000 high-quality, diverse object assemblies, providing the necessary scale and diversity to train a generalizable assembly model.

With these innovations, Assembler achieves, for the first time, reasonable 3D part assembly of general objects conditioned on a single image. Our method achieves state-of-the-art performance on the PartNet benchmark and demonstrates strong generalization to novel, complex objects. Figure~\ref{fig:teaser} shows some 3D part assembly examples of general objects. Furthermore, based on Assembler, we explore a new interesting 3D modeling system, which features part-aware high-resolution 3D object generation from a single image, and may inspire future research on high-quality part-based modeling.

Our contributions are summarized as follows: 1) We formulate 3D part assembly as a generative task via diffusion models, and introduce an effective conditioning mechanism for part-level inputs, 2) We propose a novel sparse anchor point representation for part assembly, making the generation process more scalable and semantically grounded, 3) A large-scale 3D part assembly dataset is constructed through 3D object segmentation and filtering pipeline, to facilitate the training, 4) Assembler achieves state-of-the-art results on PartNet and generalizes well to diverse real-world objects. An interesting new 3D generation system is proposed to enable part-aware, easy-to-edit, high-resolution 3D content creation. 
\section{Related Works}
\subsection{3D Part Assembly}
3D part assembly is a fundamental yet challenging task in shape modeling. Early works such as Modeling by Example~\cite{funkhouser2004modeling} tackled part re-assembly problem by building part repository, retrieving relevant parts and building low-level correspondences to assemble, followed by several works~\cite{chaudhuri2011probabilistic, kalogerakis2012probabilistic, jaiswal2016assembly} using probablistic graphical models to capture semantic and geometric relationships among shape components. While effective as an interactive user tool, these methods cannot produce fully automatic 3D part assembly. With the advent of deep neural networks and part-level 3D model annotations~\cite{mo2019partnet}, more recent works have pursued automatic part assembly. DGL~\cite{zhan2020generative} proposed an iterative graph neural network to encode part relations and predict part poses to assemble. Following this line of research, RGL~\cite{narayan2022rgl} and SPAFormer~\cite{xu2025spaformer} introduced the input part orderings to improve the assembly performance. Img-PA~\cite{li2020learning}, similar to us, relied on a single image as condition to constrain the otherwise combinatorially large assembly space. 3DHPA~\cite{du2024generative} incorporated a part-whole hierarchy to ease the difficulty of learning direct part-to-object mappings. PhysFiT~\cite{wang2024physfit} further added physical plausibility constraints to ensure that assembled objects were structurally viable.

While most of the aforementioned approaches treat part assembly as a deterministic pose prediction problem, several methods adopt a generative perspective, to better address the inherent ambiguity. Score-PA~\cite{cheng2023score} proposed to learn the probability distribution of the part poses with score-based models. DiffAssemble~\cite{scarpellini2024diffassemble} and FragmentDiff~\cite{xu2024fragmentdiff} utilized diffusion models for 3D fracture assembly problem, a highly relevant task which focus more on low-level geometric cues. We refer readers to the recent survey on 3D fragment assembly~\cite{lu2024survey} for a broader overview of this task. Our Assembler shares the similar generative motivations, and extends it with a unified, scalable diffusion model with effective encoding scheme for various input conditions such as input part meshes and images, enabling high-quality and generalizable 3D part assembly for general objects.

\subsection{3D Diffusion Models} 
Diffusion models, widely successful in 2D image and video synthesis, are rapidly advancing 3D generation. Early methods~\cite{liu2023zero, shi2023mvdream, long2024wonder3d} adapt image diffusion models for multi-view generation, while large-scale 3D datasets~\cite{deitke2023objaverse, deitke2023objaversexl} have enabled native 3D diffusion. Various representations have been explored: Point-E~\cite{nichol2022point} uses point clouds for efficiency; voxel-based models~\cite{ren2024xcube, xiong2024octfusion} employ hierarchical structures to reduce memory cost. Mesh-oriented approaches~\cite{alliegro2023polydiff, Liu2023MeshDiffusion} diffuse over mesh vertices or use deformable marching tetrahedra~\cite{shen2021deep}. Recently, 3D Gaussians~\cite{lan2024ga} offer a compact and renderable alternative.

Beyond explicit representations, implicit approaches have gained attention for their compactness, smoothness, and scalability. 3DGen \cite{gupta20233dgen} and Direct3D~\cite{wu2024direct3d} adopt triplane~\cite{chan2022efficient} representation, and 3DShape2VecSet \cite{zhang20233dshape2vecset} demonstrates the potential of latent vector sets as scalable shape embeddings, which is further advanced by works such as CraftsMan~\cite{li2024craftsman}, CLAY~\cite{zhang2024clay}, and TripoSG~\cite{li2025triposg} through larger model capacity, data scale, and compute, achieving highly detailed and diverse 3D generations. TRELLIS~\cite{xiang2024structured} proposes a hybrid sparse latent voxel representation with an enhanced 3D VAEs to better capture semantics and geometry.

These efforts collectively validate diffusion models as a scalable and expressive paradigm for 3D generation. Building on this foundation, we propose to formulate the 3D part assembly task as a diffusion-based generation of 3D anchor points, enabling both high-quality and scalable part assembly for general objects.

\subsection{Part-aware 3D Modeling}
While recent 3D generation methods can synthesize high-quality geometry from text or image prompts, they typically produce monolithic meshes without part decomposition, limiting their usefulness for editing, animation, and interaction. As a critical feature of 3D assets, part-aware 3D modeling is receiving increasing attention. The seminal work of Funkhouser \textit{et al.}~\cite{funkhouser2004modeling} created 3D shapes by part retrieving and re-assembly, followed by subsequent efforts~\cite{chaudhuri2011probabilistic, kalogerakis2012probabilistic} to improve both the geometry and semantic fidelity. With the advent of deep learning, several methods leverage neural networks to learn part-aware 3D modeling including Shape VAE~\cite{nash2017shape},  per-part VAE-GANs~\cite{li2020learning}, or Seq2Seq networks~\cite{wu2020pqnet}. These approaches demonstrated the feasibility of decomposed 3D generation, though often limited to specific categories and coarse outputs. More recently, Part123~\cite{liu2024part123} and PartGen~\cite{chen2024partgen} advanced part-aware generation by first generating and segmenting 2D multi-view images, then lifting them into 3D using multiview reconstruction. Complementarily, HoloPart~\cite{yang2025holopart} generates full 3D objects and then performs 3D segmentation and part completion, improving the geometric quality of individual components.

Our Assembler provides a solid base for part-aware 3D modeling. Based on it, we introduce a interesting neural-symbolic pipeline that integrates top-down reasoning via vision-language models (VLMs), 3D part generation, and bottom-up part assembly. It offers a new symbolic perspective for part-aware, high-resolution 3D modeling.
\begin{figure*}
    \centering
    \includegraphics[width=1.0\linewidth]{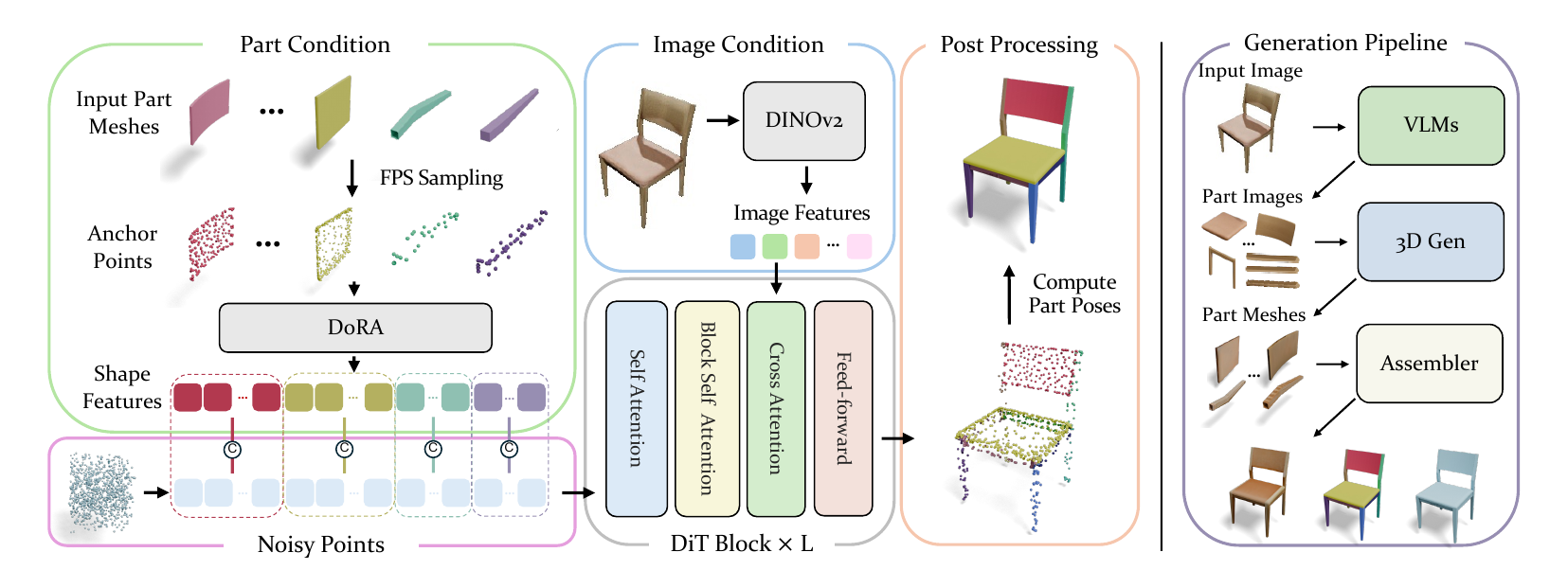}
    \vspace{-20pt}
    \caption{Overview of Assembler (Left) and part-aware 3D generation pipeline (Right). (Left) The input part meshes are sampled as anchor points representation, followed by DoRA to extract shape features. These shape features are concatenated with noised point tokens, and a diffusion model is trained to generate assembled anchor points. After that, a simple least-squares fitting is used to compute part poses from generated and input anchor points to assemble the input meshes as a final object. (Right) The input image is first fed into VLMs to infer the parts and generate reference images for each part. Then an image-to-3D generator is applied to produce part meshes. Finally, Assembler generates complete, high-resolution, part-aware 3D models by assembling the part meshes.}
    \label{fig:pipeline}
\end{figure*}
\section{Method}
Assembler formulates the 3D part assembly as a generative task, and proposes a novel diffusion model with sparse anchor point representation, enabling scalable part assembly for general objects. The overall framework is illustrated in Figure~\ref{fig:pipeline}. Next, we will describe in detail the assembly representation (Sec.~\ref{sec:representation}), the assembly diffusion model (Sec.~\ref{sec:diffusion}), the curation of assembly data (Sec.~\ref{sec:data}), and a prototype of part-aware 3D generation pipeline (Sec.~\ref{sec:application}).

\subsection{Assembly Representation}
\label{sec:representation}
Given a set of $N$ input parts $\mathcal{P} = \{\mathcal{P}_1,..., \mathcal{P}_N\}$ in mesh or point cloud formats, the 3D part assembly task aims to generate a complete shape $\mathcal{S} = \bigcup_i^N\mathcal{P}_i^\prime$, where $\mathcal{P}_i^\prime=\mathcal{T}(\mathcal{P}_i, \mathrm{T}_i)$ indicates the transformed part using rigid transformation $\mathrm{T}_i$. It is then straight-forward to design deterministic or generative networks to predict the per-part transformation $\mathrm{T}_i$ to assemble the parts, as done in previous works. However, directly learning the SE(3) manifold distribution is not as intuitive or easy as learning in Euclidean space, often causing unstable or suboptimal performance~\cite{leach2022denoising}. The resulting pose space describes the underlying distribution of part poses, which is hardly semantic by itself, and relies on the input parts to form as a meaningful sample. This hinders the learning of a compact, smooth, continuous latent space for part assembly. 

To tackle this problem, we propose a sparse anchor point representation for 3D part assembly. Specifically, each part $\mathcal{P}_i$ is represented by a sparse point cloud $\mathbf{p}_i$ sampled from its original surface or dense points. The part assembly task is then defined as the generation of a complete object point cloud $\mathcal{X}=\bigcup_i^N \mathbf{p}^\prime_i$, where $\mathbf{p}_i^\prime$ is the generated transformed anchor points for each part. In other words, instead of generating each part pose, we directly generate the resulting sparse anchor point clouds of each part in the shared global object coordinates, as shown in Figure~\ref{fig:pipeline}. After acquiring the assembled object points, we perform a simple least-square-based fitting to compute the transformation between the initial and generated anchor points. This per-part transformation $\mathrm{T}_i$ is then used to transform the input mesh or dense point clouds to assemble the final object.

With our proposed anchor point representation, the generative model now directly learns the object-level shape distributions in Euclidean space with rich semantics, enabling scalable and high-quality generation. Moreover, the anchor point representation is flexible to handle the varying number of parts in both training and inference. We control the same total number of $M=M_1+\cdots+M_N$ anchor points for the assembled objects, while adaptively assigning different numbers of anchor points on each part, according to the part numbers and their sizes. With only $M=256$ or $1024$ anchor points for an object, it theoretically supports hundreds of parts, since each part at minimum requires two anchor points to compute the transformation. This flexibility greatly facilitates the scalable generation of part assemblies.

Despite its strengths, generating transformed anchor points is not trivial. The key challenges include effective part-level information encoding with sparse anchor points, maintaining rigid part transformations, and preserving the same point ordering within each part. To handle these issues, we design a dedicated assembly diffusion model for generating high-quality assembled object points under input part conditions.

\subsection{Assembly Diffusion Model}
\label{sec:diffusion}
Our proposed assembly diffusion model aims to sample a complete object point cloud $\mathcal{X}$, which ideally should be the concatenation of all rigidly transformed sparse anchor points as $[p_1^\prime,...,p_N^\prime]$. This corresponds directly to the input anchor points $[p_1,...,p_N]$, preserving both part order and intra-part point ordering. Each generated part $p_i^\prime$ should retain the geometric structure of the original $p_i$, simulating a rigid transformation. To ensure these properties, we introduce several key design choices for the assembly diffusion model.

\noindent\textbf{Model Architecture.} We employ the popular Diffusion Transformer (DiT) model as the denoise function $\epsilon_{\theta}$, to predict the noise at each timestamp $t$ via $L$ layers of stacked cross and self attentions:
\begin{equation}
    \epsilon_{\theta}(\mathbf{x_t}, t) = \{\operatorname{CrossAttn}(\operatorname{SelfAttn}(\operatorname{SelfAttn}(\mathbf{x_t\#\mathbf{c}_p}), \mathbf{m}), \mathbf{c_g})\}^{L}
    \label{eq::dit}
\end{equation}
where $\mathbf{x_t}$ is the noisy version of clean data $\mathbf{x_0}$, and $\mathbf{c_p}, \mathbf{c_g}$ are the part and image conditions, respectively. $\#$ indicates concatenation. Certain feed-forward layers are omitted from this description for simplicity. Besides cross-attention to aggregate features and self-attention to encode globally, we introduce an additional self-attention layer to encode the per-part information within each part. This is achieved via a block diagonal matrix $\mathbf{m}$ as the attention mask, restricting the tokens only attends to those belonging to the same part. The DiT model is capable of encoding intra-part shape information, reasoning about inter-part relations, and learning the complete shape priors, which are all essential for a good 3D part assembly.

\noindent\textbf{Condition Scheme.} We treat input parts and the reference image as diffusion conditions, and design effective condition schemes for them respectively. For the reference image, we extract the spatial patch features $\mathbf{c_g}$ with pretrained DINOv2~\cite{oquab2023dinov2} model, and inject into the model with cross-attention. 

For the input part meshes, instead of using cross-attention, we directly concatenate part condition $\mathbf{c_p}\in\mathbb{R}^{M\times C}$ with the input noise $\mathbf{x_t}\in\mathbb{R}^{M\times 3}$, to enable point-aligned interactions between all anchor points and the noised assembled points, which keeps at best the rigidity and point ordering of input anchor points in the final generated object point cloud. Specifically, each part condition consists of three components, the original sparse anchor point coordinates $p_i$, the part shape latents $d_i$, and part index embedding $e_i$. For each part, we first sample a dense point cloud $p_i^d\in\mathbb{R}^{Q\times 3}$ to approximate the surface geometry, and then sample the sparse anchor points $p_i\in\mathbb{R}^{M_i\times 3}$ from the dense point cloud. Note that the number $M_i$ of sparse anchor points for each part is determined adaptively, to ensure the fixed total number of $M$ points for generation. Using only sparse anchor point coordinates is obviously not sufficient, thus we use the pre-trained Dora VAE~\cite{chen2024dora}, a 3D shape variational auto-encoder, to encode the part geometry into sparse anchor points as shape latents. We feed the dense point cloud $p_i^d$ and the sparse anchor points $p_i$ to the Dora encoder as shape points and queries, resulting in $d_i\in\mathbb{R}^{M_i\times C_d}$ anchor point latent features. 
Furthermore, to eliminate the ambiguity of repetitive parts, we introduce part index embedding $e_i\in\mathbb{R}^{M_i\times C_e}$ as the Fourier positional embedding of the part index. All three components $p_i, d_i, e_i$ are concatenated to form the condition $c_i \in \mathbb{R}^{M_i\times (3+C_d+C_e)}$ for each part. Together all input parts can be concatenated into the above part condition $\mathbf{c_p}\in\mathbb{R}^{M\times (3+C_d+C_e)}$, which is then concatenated with the noise for aligned conditioning, as shown in Eq.~\ref{eq::dit} and Figure~\ref{fig:pipeline}.

\noindent\textbf{Post Processing.} Our ultimate goal is to assemble the input part meshes, not only the sparse anchor points. Thus, we need to compute the transformation for each part mesh from the generated anchor points $\mathcal{X}$. To do so, we can simply fetch the corresponding $p_i^\prime$ of the input part $p_i$, and use least square fitting to calculate the transformation. The final assembly is then acquired by transforming every input part mesh in this way.

\subsection{Data Curation}
\label{sec:data}
A large-scale, high-quality, diverse 3D part assembly dataset is essential for scaling up assembly model to general objects. Unfortunately, such a dataset is not available now. To fill this gap, we curate a large-scale assembly dataset by collecting and synthesizing 3D part assembly from existing data resources. A simple yet effective assembly data synthesis pipeline is proposed to create part assembly data from 3D complete meshes, including data filtering, segmentation, grouping, and augmentation. 

Specifically, the 3D mesh data is first filtered to remove complex 3D scene data and low-quality scanned meshes. Then, we segment the mesh into parts. While state-of-the-art 3D part segmentation methods~\cite{yang2024sampart3d, zhou2023partslip++} can be utilized, they still suffer from inescapable errors and long processing time. In contrast, we observe that a large portion of 3D meshes in existing 3D datasets~\cite{deitke2023objaverse} are created by artists, and these meshes can be easily split into parts by only checking the connected components of faces. Meshes that cannot be split in this way, or meshes that contain a single dominant part after splitting, are then filtered out. This segmentation is fast and high-quality, preserving all the geometric units of the original meshes. However, the resulting parts are often too small and over-segmented in terms of semantic parts. Thus, we next perform grouping on adjacent geometric parts to better represent semantics. A simple KNN on part centroids is applied to group the parts into desired part number, which is randomly choosed from 3 to 100 parts. Finally, with these segmented parts, we apply random rotation and translation to each part, to simulate the starting poses for part assembly.

Leveraging above data synthesizing pipeline, we process the TRELLIS-500K dataset~\cite{xiang2024structured} filtered from Objaverse-XL~\cite{deitke2023objaverse}, ABO~\cite{collins2022abo}, 3D-Future~\cite{fu20213d}, HSSD~\cite{khanna2024habitat} and Toys4K~\cite{stojanov2021using}. Furthermore, we also process ShapeNet~\cite{chang2015shapenet} as a complement. Together with existing part dataset~\cite{mo2019partnet}, we curate in total around 320K objects with their part assemblies. The detailed statistics and examples are shown in supplementary materials. This dataset and its curation method not only provide a solid foundation for training general 3D part assembly model, but also could facilitate broader part-based 3D modeling tasks.

\begin{figure*}[tb]
    \centering
    \small
    \setlength{\tabcolsep}{0pt}
    \begin{tabular}{ccccccc}

    \centering
        {\includegraphics[width=0.13\linewidth]{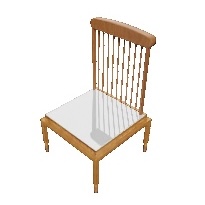}} &
        {\includegraphics[width=0.13\linewidth]{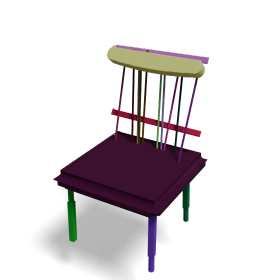}} & 
        {\includegraphics[width=0.13\linewidth]{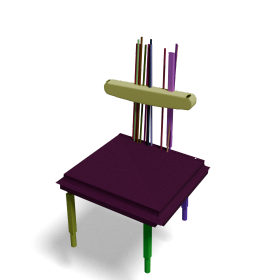}} &
        {\includegraphics[width=0.13\linewidth]{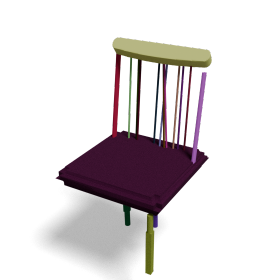}} &
        {\includegraphics[width=0.13\linewidth]{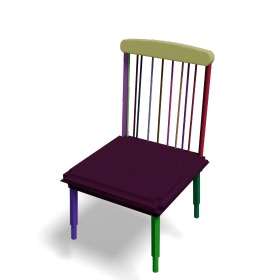}} &
        {\includegraphics[width=0.13\linewidth]{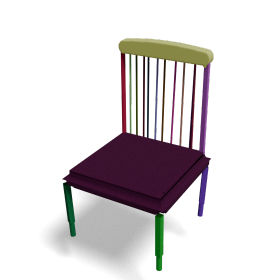}} &
        {\includegraphics[width=0.13\linewidth]{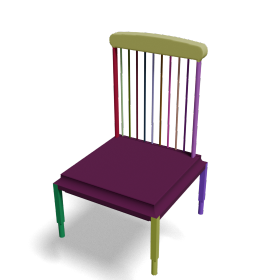}} \\

        {\includegraphics[width=0.12\linewidth]{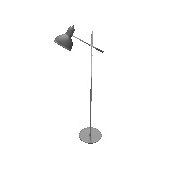}} &
        {\includegraphics[width=0.12\linewidth]{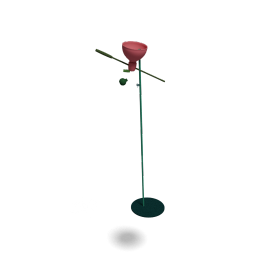}} &
        {\includegraphics[width=0.12\linewidth]{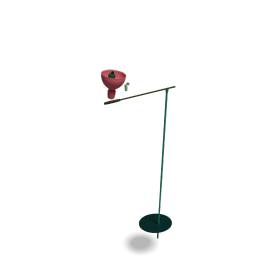}} &
        {\includegraphics[width=0.12\linewidth]{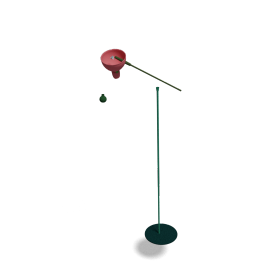}} &
        {\includegraphics[width=0.12\linewidth]{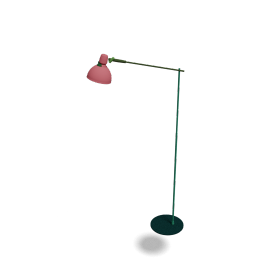}} &
        {\includegraphics[width=0.12\linewidth]{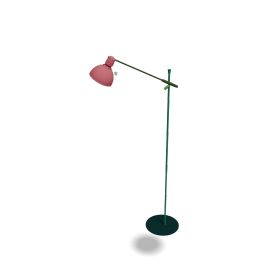}} &
        {\includegraphics[width=0.12\linewidth]{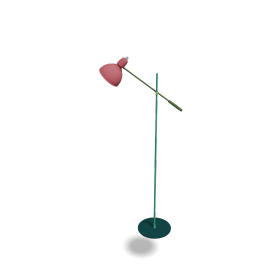}} \\

        Reference Image & DGL & Score-PA & SPAFormer & Ours & Ours-img & Groundtruth \\

    \end{tabular}
    \caption{Qualitative comparison of category-specific 3D part assembly on PartNet dataset.}
    \label{fig:partnet}
\end{figure*}
\begin{table*}
\centering
\caption{Quantitative comparison of category-specific models on PartNet dataset. \textbf{Best}, \underline{second best} results are highlighted.}
\vspace{-5pt}
\setlength{\tabcolsep}{3pt}
\scalebox{0.9}{
\begin{tabular}{l|cccc|cccc|cccc}
\toprule
\centering
&\multicolumn{4}{c|}{Chair}& 
\multicolumn{4}{c|}{Table}&
\multicolumn{4}{c}{Lamp} \\
\midrule
Method & SCD ↓ & PA ↑ & CA ↑ & SR ↑ & SCD ↓ & PA ↑ & CA ↑ & SR ↑ & SCD ↓ & PA ↑ & CA ↑ & SR ↑  \\
\midrule 
DGL~\cite{zhan2020generative} &  9.60 & 37.50 & 24.26 & 7.60 & - & - &
- & - & \textbf{7.82} & 32.33 & 42.06 & 12.38 \\
RGL~\cite{narayan2022rgl} & 10.23 & 43.82 & 25.97 & 7.43 & - & - & - & - & 15.24 & 34.19 & 47.18 & 14.86 \\ 
Score-PA~\cite{cheng2023score} & 7.41 & 41.74 & 28.38 & 5.06 & \underline{4.54} & 51.57 &
39.85 & 7.53 & \underline{8.68} & 31.05 & 48.12 & 8.59 \\
SPAFormer~\cite{xu2025spaformer} & \textbf{7.02} & 55.48 & 39.16 & 13.68 & \textbf{3.93} & 64.14 & \textbf{61.90} & 33.98 & 12.80 & 36.76 & 45.66 & 15.54 \\
\midrule
Ours &  9.20 & \underline{61.85} & \underline{49.82} & \underline{22.59} & 8.02 & \underline{65.04} & 56.15 & \underline{39.07} & 26.88 & \underline{44.04} & \textbf{58.16} & \underline{27.88} \\
Ours-img & \underline{7.21} & \textbf{66.17} & \textbf{52.20} & \textbf{24.92} & 4.72 & \textbf{70.44} &
\underline{57.72}  & \textbf{43.46} & 16.24  & \textbf{51.42} & \underline{57.44} & \textbf{33.24} \\
\bottomrule
\end{tabular}}
\label{tab:partnet}
\end{table*}
\subsection{Part-aware 3D Generation Pipeline}
\label{sec:application}
With above proposed representation, model and data, we can now train a scalable 3D part assembly model, for the first time achieving part assembly of general objects, which enables exciting downstream applications. Here we introduce a new part-aware 3D generation prototype system, based on the symbolic calls of large Vision-Language Models (VLM), 3D generator, and our Assembler.

Given a image as input, existing 3D generation methods~\cite{li2025triposg, zhang2024clay, li2024craftsman} mostly create a single monolithic mesh without separated parts, and its 3D geometric details or 3D "resolution" are bounded by the capability of the model, which limits their quality and usability. In contrast, artists often create 3D assets by modeling each part or reusing existing part assets, and assembling them into a complete model. Inspired by this, we propose a system mimicking this pipeline. Given an object image, we first use GPT-4o~\cite{hurst2024gpt} to analyze the image and identify all the semantic parts of the object. Then, we ask GPT-4o to generate a high-resolution reference image for each part according to the original image. After getting the images for each part, we input these images to a image-to-3D generator like TripoSG~\cite{li2025triposg}, to generate the 3D shape of each part. Finally, all the generated parts are assembled by our Assembler as a part-aware, high-resolution 3D model. In this framework, the reasoning, segmentation, and completion of object parts are handed to the VLM which has rich 2D priors. The 3D generator then focus on the high-resolution-part-image-to-3D task with iterative calls to eliminate the "3D resolution" limit. Finally, our Assembler is essential to automatically bring these 3D parts into a complete 3D model. More details are provided in supplementary materials. This innovative part-aware 3D generation pipeline demonstrates new potentials for 3D content creation towards high-quality and easy-to-use 3D models.
\section{Experiments}
\subsection{Implementation Details}
For the total anchor point number $M$, we use 1024 to balance efficiency and accuracy. The numbers of anchor points for each part are adaptively assigned based on their size ratios. The dense point sampling number $Q$ for each part is set to 4096. For the DiT architecture, we train a relatively small model (49M parameters) for category-specific comparison with baselines. A large model (195M) is trained for general objects on our constructed dataset. We refer to supplementary material for more implementation details.

\subsection{Category-specific Comparison}
Since previous assembly methods mostly trained and evaluated category-specific models on PartNet~\cite{mo2019partnet}, we follow their settings and train three specific models for Chair, Table and Lamp. For evaluation, we follow existing works and use Shape Chamfer Distance (SCD), Part Accuracy (PA), Connectivity Accuracy (CA), and Success Rate (SR) as metrics. SCD measures the chamfer distance between the assembled and groundtruth shape, and PA counts parts placed within a 0.01 distance threshold. CA checks the correctness of adjacent part connections. SR denotes the percentage of successful assemblies where all parts meet the PA threshold.

\begin{figure*}[tb]
    \centering
    \small
    \setlength{\tabcolsep}{0pt}
    \begin{tabular}{ccccccc}

    \centering    
        {\includegraphics[width=0.13\linewidth]{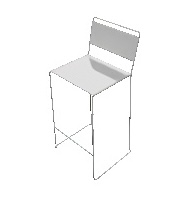}} &
        {\includegraphics[width=0.13\linewidth]{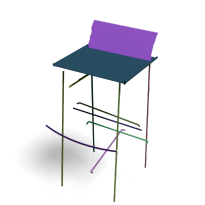}} &
        {\includegraphics[width=0.13\linewidth]{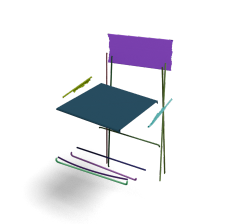}} &
        {\includegraphics[width=0.13\linewidth]{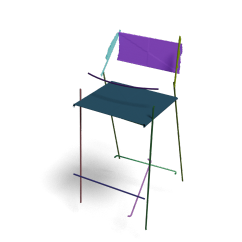}} &
        {\includegraphics[width=0.13\linewidth]{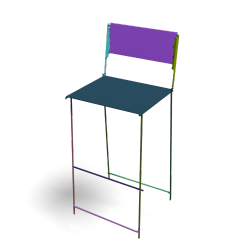}} &
        {\includegraphics[width=0.13\linewidth]{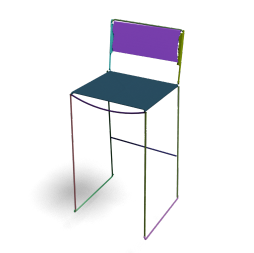}} &
        {\includegraphics[width=0.13\linewidth]{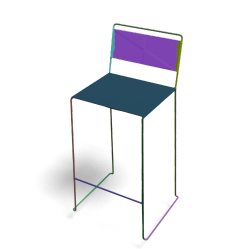}} \\

        {\includegraphics[width=0.13\linewidth]{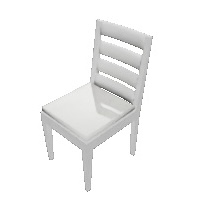}} &
        {\includegraphics[width=0.13\linewidth]{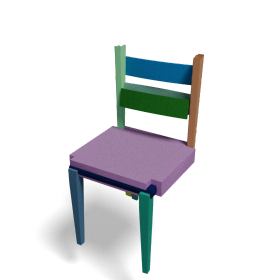}} &
        {\includegraphics[width=0.13\linewidth]{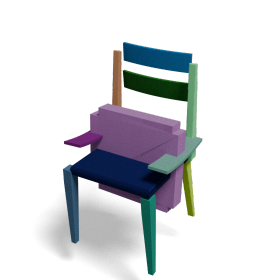}} &
        {\includegraphics[width=0.13\linewidth]{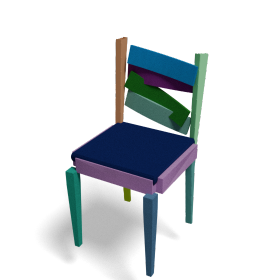}} &
        {\includegraphics[width=0.13\linewidth]{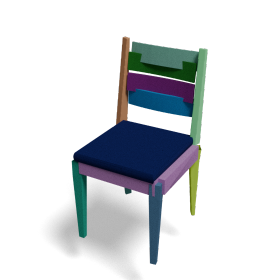}} &
        {\includegraphics[width=0.13\linewidth]{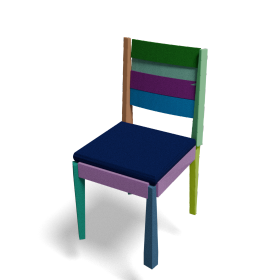}} &
        {\includegraphics[width=0.13\linewidth]{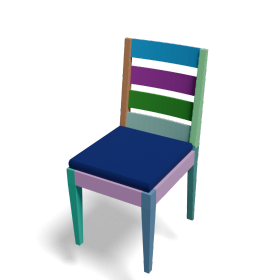}} \\

        {\includegraphics[width=0.13\linewidth]{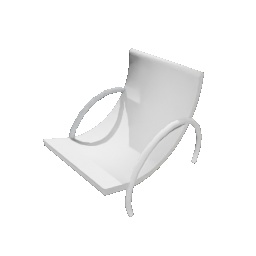}} &
        {\includegraphics[width=0.13\linewidth]{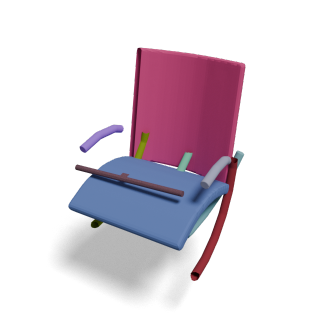}} &
        {\includegraphics[width=0.13\linewidth]{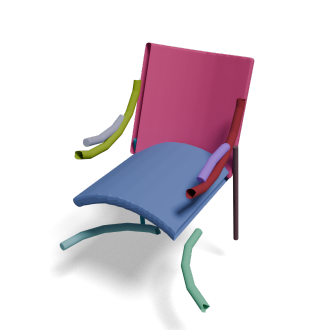}} &
        {\includegraphics[width=0.13\linewidth]{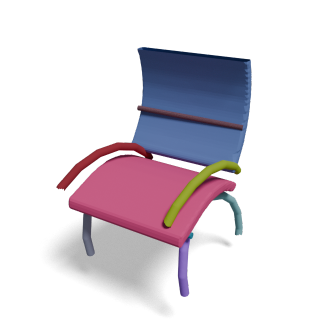}} &
        {\includegraphics[width=0.13\linewidth]{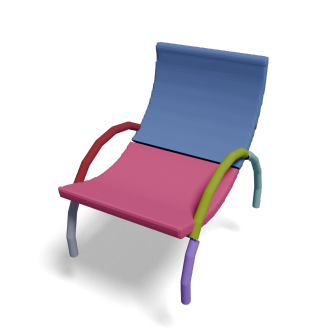}} &
        {\includegraphics[width=0.13\linewidth]{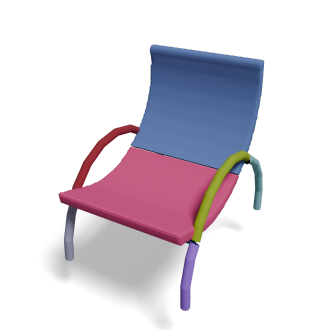}} &
        {\includegraphics[width=0.13\linewidth]{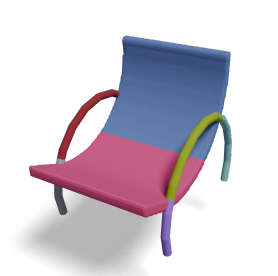}} \\


        {\includegraphics[width=0.13\linewidth]{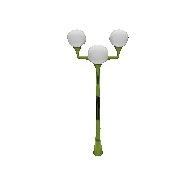}} &
        {\includegraphics[width=0.13\linewidth]{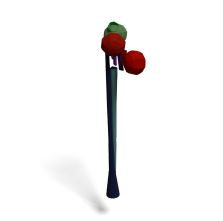}} &
        {\includegraphics[width=0.13\linewidth]{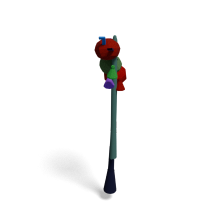}} &
        {\includegraphics[width=0.13\linewidth]{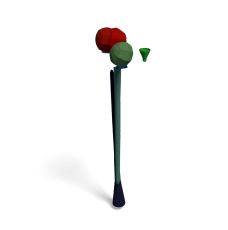}} &
        {\includegraphics[width=0.13\linewidth]{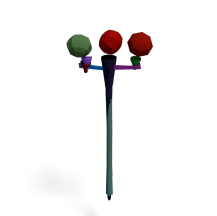}} &
        {\includegraphics[width=0.13\linewidth]{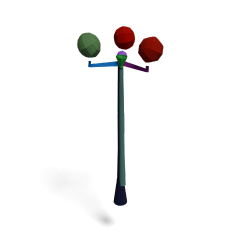}} &
        {\includegraphics[width=0.13\linewidth]{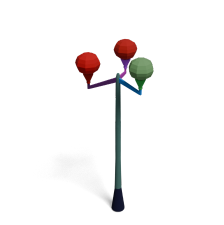}} \\

        {\includegraphics[width=0.13\linewidth]{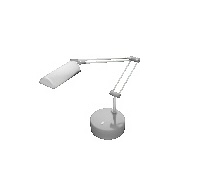}} &
        {\includegraphics[width=0.13\linewidth]{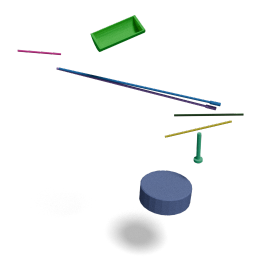}} &
        {\includegraphics[width=0.13\linewidth]{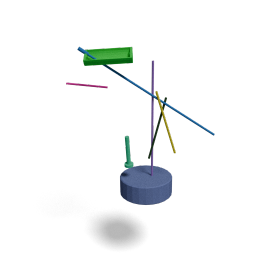}} &
        {\includegraphics[width=0.13\linewidth]{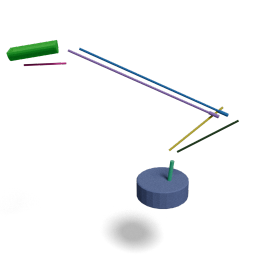}} &
        {\includegraphics[width=0.13\linewidth]{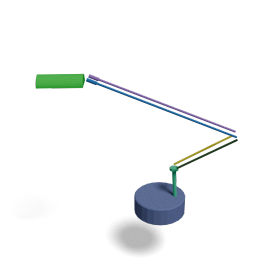}} &
        {\includegraphics[width=0.13\linewidth]{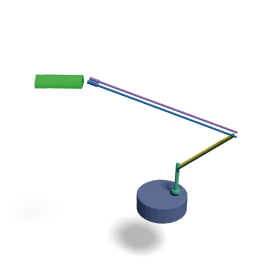}} &
        {\includegraphics[width=0.13\linewidth]{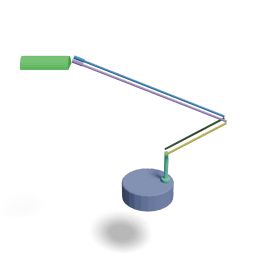}} \\

        Reference Image & DGL & Score-PA & SPAFormer & Ours & Ours-img & Groundtruth \\

    \end{tabular}
    \caption{More qualitative comparison of category-specific 3D part assembly on PartNet dataset.}
    \Description{More qualitative comparison of category-specific 3D part assembly on PartNet dataset.}
    \label{fig_only:partnet}
\end{figure*}

We compare with four open-source representative methods. For a fair and comprehensive comparison, we include two versions of our models: \textit{Ours} and \textit{Ours-img} indicates the DiT models without and with image condition, respectively. As shown in Table~\ref{tab:partnet}, our model significantly outperforms baselines across all three categories, particularly in PA and SR, indicating a stronger ability to successfully and accurately assemble objects. While our performance on the SCD metric is slightly lower, this is likely due to the inherent ambiguity in part assemblies—our model may produce plausible configurations that differ from the ground truth. Unlike PA, CA, and SR, which tolerate such variations to some extent, SCD penalizes them more heavily as it measures geometric distances rather than correctness. Adding an image condition to our model partially relieves this ambiguity issue and results in a consistently better performance. We show qualitative comparison in Figure~\ref{fig:partnet} and Figure~\ref{fig_only:partnet}. While the baselines suffer from thin structures, repetitive parts, and complex assemblies, our models can effectively handle these and generate accurate and valid assemblies.

\begin{figure*}[tb]
    \centering
    \small
    \setlength{\tabcolsep}{0pt}
    \begin{tabular}{cccccccc}

    \centering
        {\includegraphics[width=0.12\linewidth]{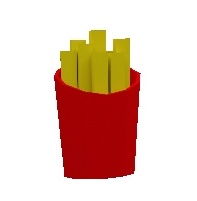}} & 
        {\includegraphics[width=0.12\linewidth]{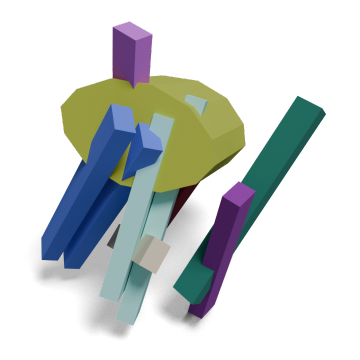}} & 
        {\includegraphics[width=0.12\linewidth]{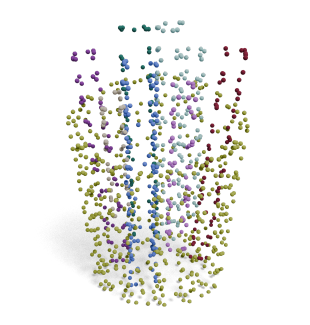}} & 
        {\includegraphics[width=0.12\linewidth]{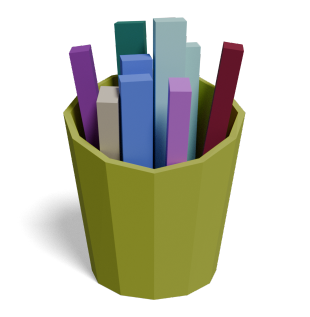}} & 

        {\includegraphics[width=0.12\linewidth]{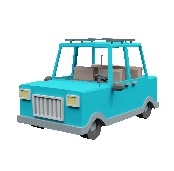}} & 
        {\includegraphics[width=0.12\linewidth]{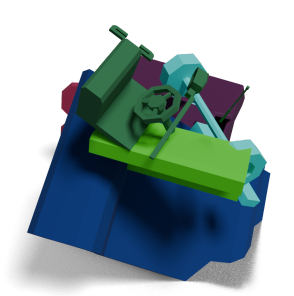}} &
        {\includegraphics[width=0.12\linewidth]{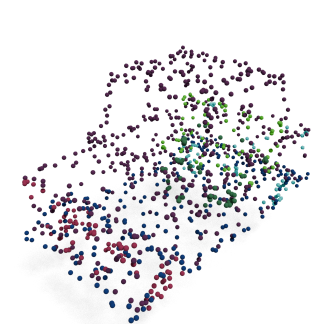}} &
        {\includegraphics[width=0.12\linewidth]{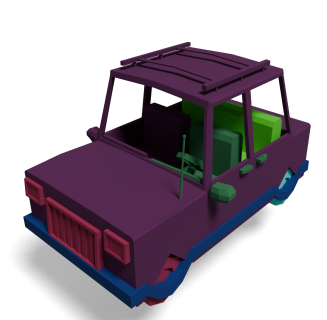}} \\

        {\includegraphics[width=0.12\linewidth]{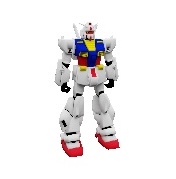}} &
        {\includegraphics[width=0.12\linewidth]{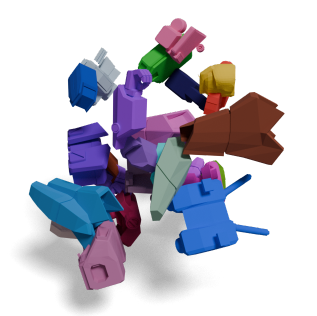}} &
        {\includegraphics[width=0.12\linewidth]{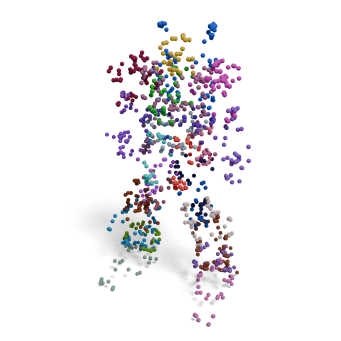}} &
        {\includegraphics[width=0.12\linewidth]{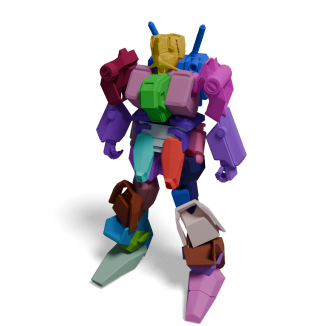}} &

        {\includegraphics[width=0.12\linewidth]{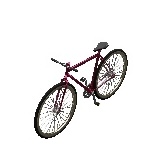}} &
        {\includegraphics[width=0.12\linewidth]{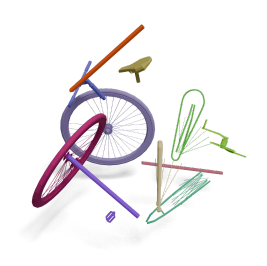}} &
        {\includegraphics[width=0.12\linewidth]{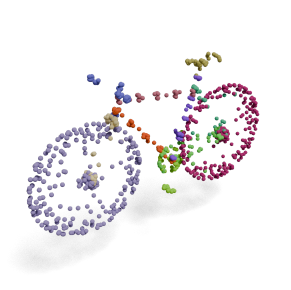}} &
        {\includegraphics[width=0.12\linewidth]{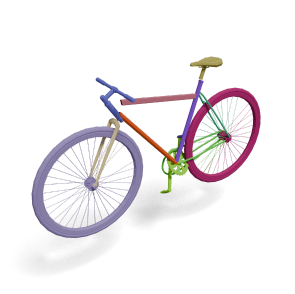}} \\

        {\includegraphics[width=0.12\linewidth]{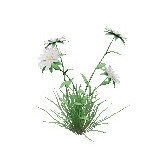}} &
        {\includegraphics[width=0.12\linewidth]{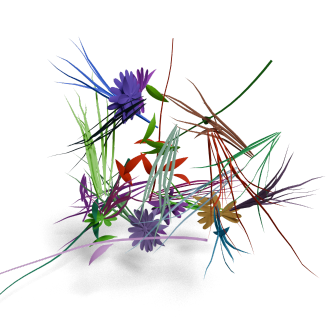}} &
        {\includegraphics[width=0.12\linewidth]{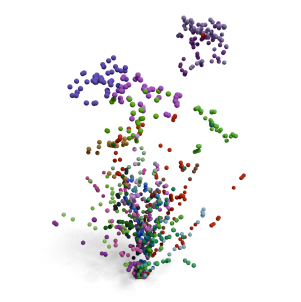}} &
        {\includegraphics[width=0.12\linewidth]{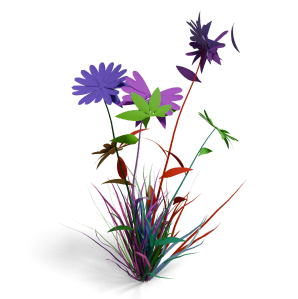}} &

        {\includegraphics[width=0.12\linewidth]{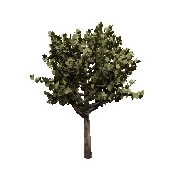}} &
        {\includegraphics[width=0.12\linewidth]{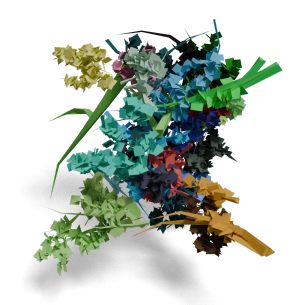}} &
        {\includegraphics[width=0.12\linewidth]{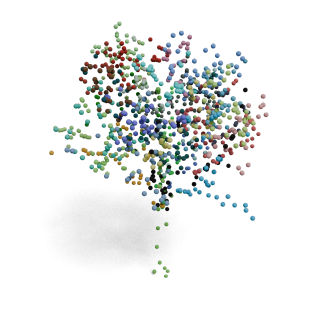}} &
        {\includegraphics[width=0.12\linewidth]{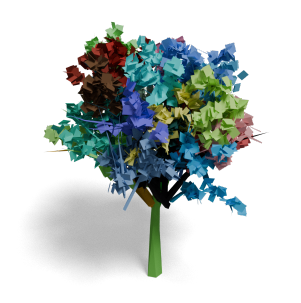}} \\

        Input Image & Input Parts & Anchor Points & Assembly & Input Image & Input Parts & Anchor Points & Assembly \\

    \end{tabular}
    \caption{3D part assembly results on Toys4K dataset. Given an input image and parts, Assembler generates anchor points and then computes the assembly.}
    \label{fig:toys4k}
    \vspace{-3pt}
\end{figure*}
\subsection{General Object Assembly Results}
\begin{figure*}[t]
    \centering
    \small
    \setlength{\tabcolsep}{0pt}
    \begin{tabular}{cccccccc}

        \centering
        {\includegraphics[width=0.12\linewidth]{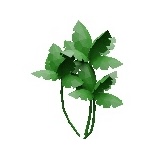}} &
        {\includegraphics[width=0.12\linewidth]{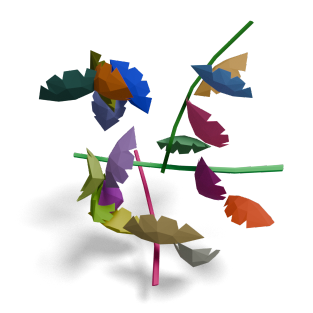}} &
        {\includegraphics[width=0.12\linewidth]{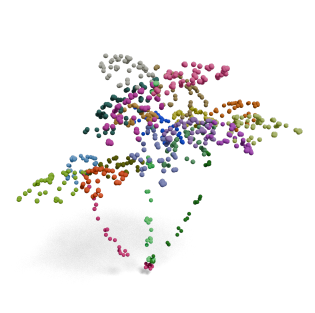}} &
        {\includegraphics[width=0.12\linewidth]{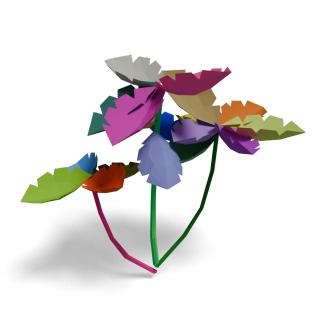}} &

        {\includegraphics[width=0.12\linewidth]{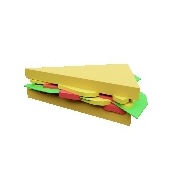}} &
        {\includegraphics[width=0.12\linewidth]{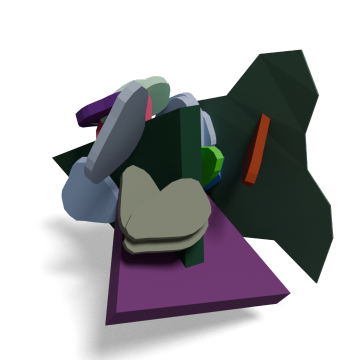}} &
        {\includegraphics[width=0.12\linewidth]{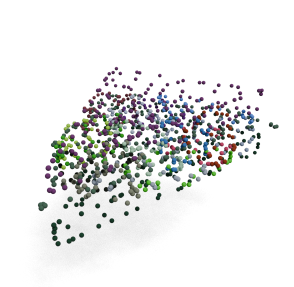}} &
        {\includegraphics[width=0.12\linewidth]{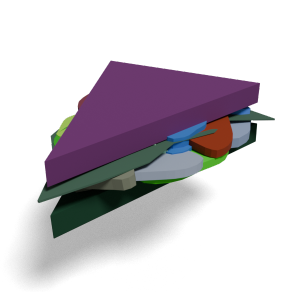}} \\

        {\includegraphics[width=0.12\linewidth]{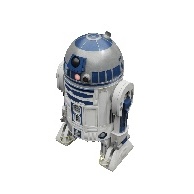}} & 
        {\includegraphics[width=0.12\linewidth]{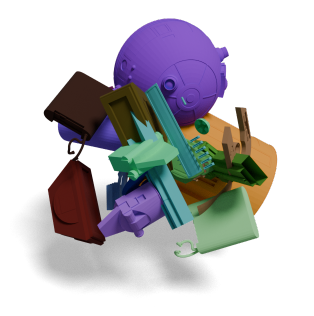}} &
        {\includegraphics[width=0.12\linewidth]{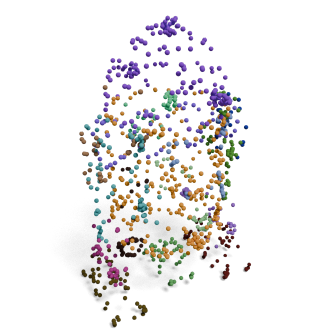}} &
        {\includegraphics[width=0.12\linewidth]{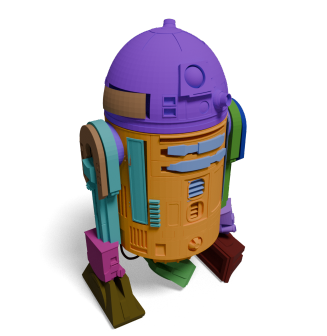}} &

        {\includegraphics[width=0.12\linewidth]{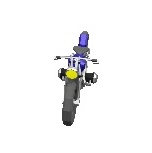}} & 
        {\includegraphics[width=0.12\linewidth]{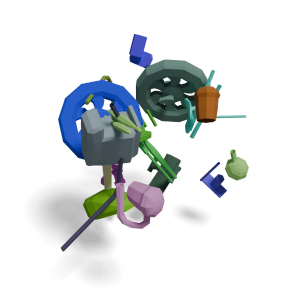}} & 
        {\includegraphics[width=0.12\linewidth]{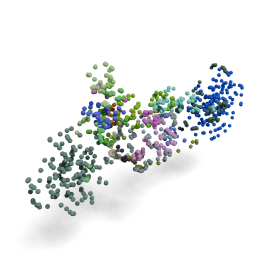}} & 
        {\includegraphics[width=0.12\linewidth]{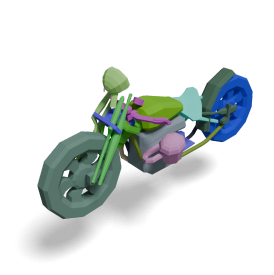}} \\

        {\includegraphics[width=0.12\linewidth]{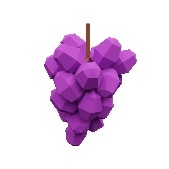}} & 
        {\includegraphics[width=0.12\linewidth]{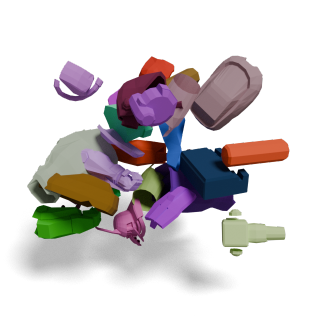}} & 
        {\includegraphics[width=0.12\linewidth]{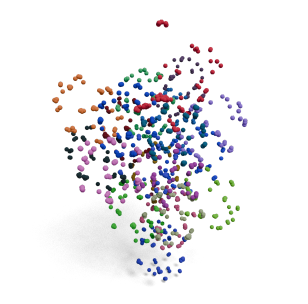}} & 
        {\includegraphics[width=0.12\linewidth]{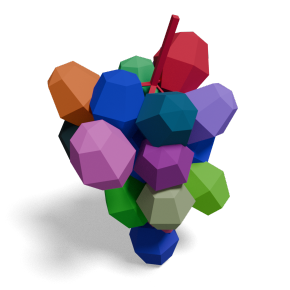}} & 

        {\includegraphics[width=0.12\linewidth]{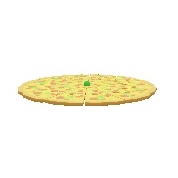}} & 
        {\includegraphics[width=0.12\linewidth]{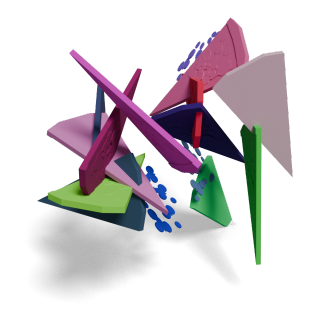}} & 
        {\includegraphics[width=0.12\linewidth]{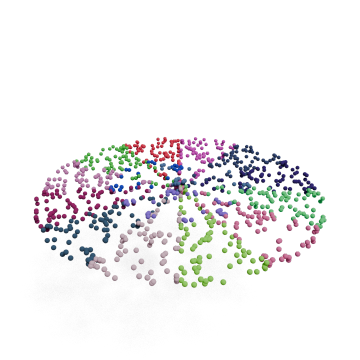}} & 
        {\includegraphics[width=0.12\linewidth]{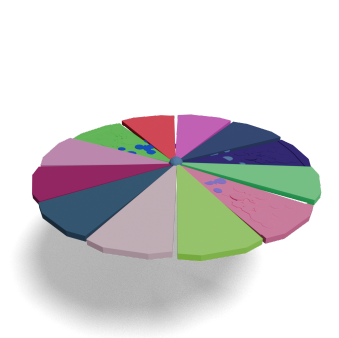}} \\

        {\includegraphics[width=0.12\linewidth]{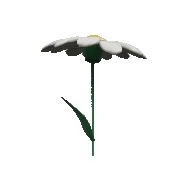}} & 
        {\includegraphics[width=0.12\linewidth]{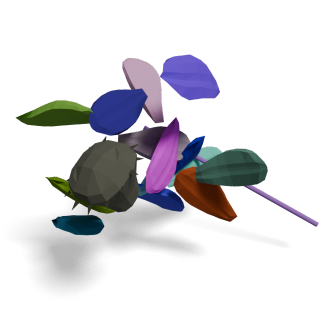}} & 
        {\includegraphics[width=0.12\linewidth]{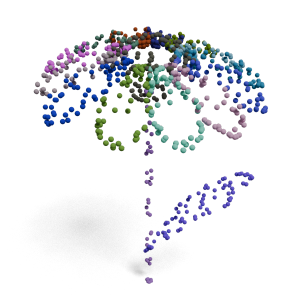}} & 
        {\includegraphics[width=0.12\linewidth]{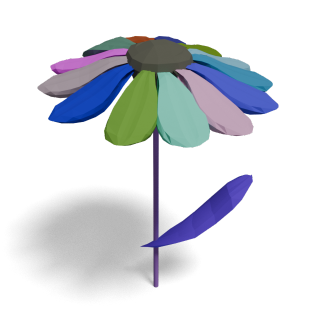}} & 

        {\includegraphics[width=0.12\linewidth]{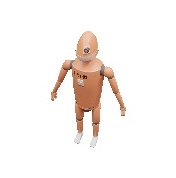}} & 
        {\includegraphics[width=0.12\linewidth]{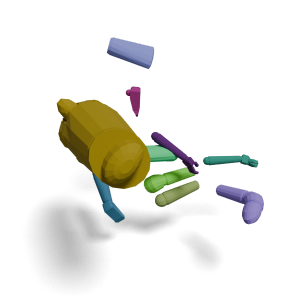}} & 
        {\includegraphics[width=0.12\linewidth]{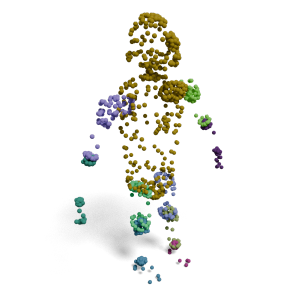}} & 
        {\includegraphics[width=0.12\linewidth]{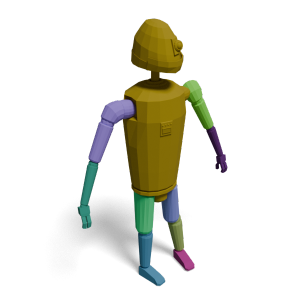}} \\

        {\includegraphics[width=0.12\linewidth]{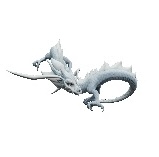}} &
        {\includegraphics[width=0.12\linewidth]{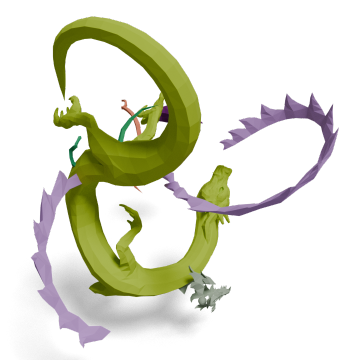}} &
        {\includegraphics[width=0.12\linewidth]{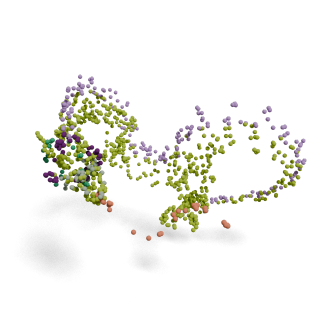}} &
        {\includegraphics[width=0.12\linewidth]{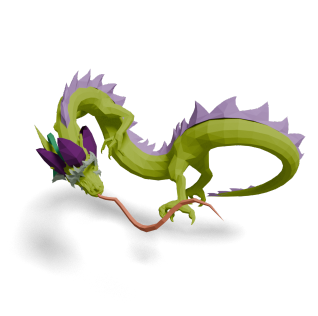}} &

        {\includegraphics[width=0.12\linewidth]{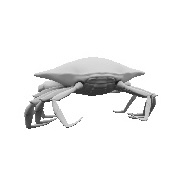}} &
        {\includegraphics[width=0.12\linewidth]{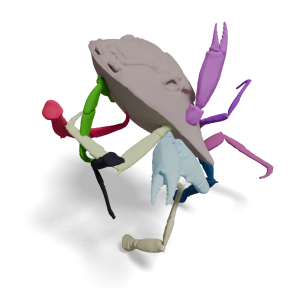}} & 
        {\includegraphics[width=0.12\linewidth]{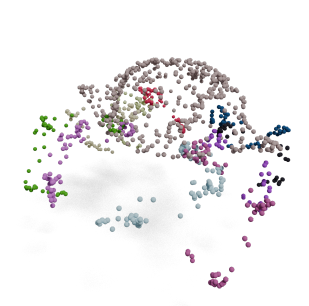}} &
        {\includegraphics[width=0.12\linewidth]{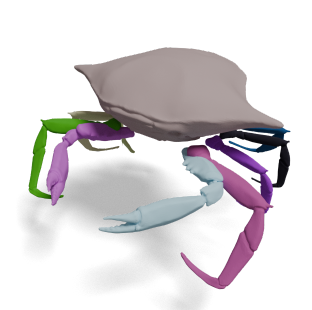}} \\

        {\includegraphics[width=0.12\linewidth]{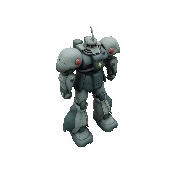}} &
        {\includegraphics[width=0.12\linewidth]{figures/images/toys4k_resize/2a297c_019_init.png}} &
        {\includegraphics[width=0.12\linewidth]{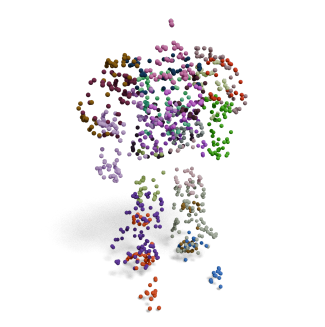}} &
        {\includegraphics[width=0.12\linewidth]{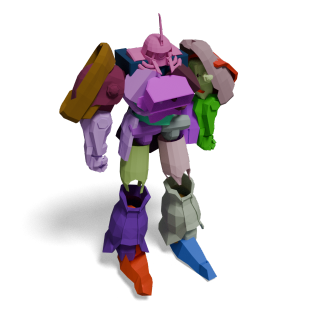}} &

        {\includegraphics[width=0.12\linewidth]{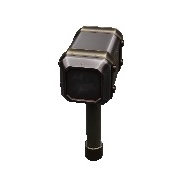}} &
        {\includegraphics[width=0.12\linewidth]{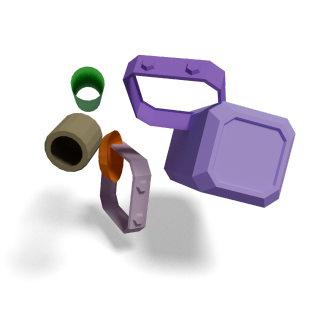}} &
        {\includegraphics[width=0.12\linewidth]{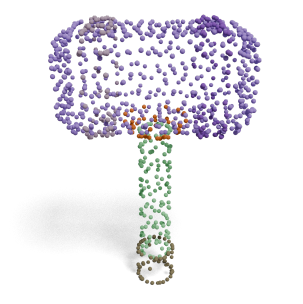}} &
        {\includegraphics[width=0.12\linewidth]{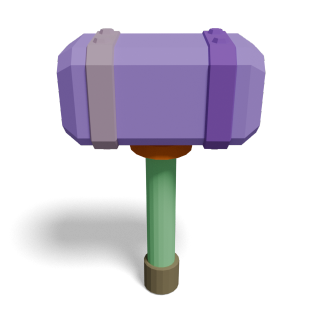}} \\

        {\includegraphics[width=0.12\linewidth]{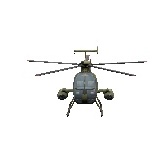}} &
        {\includegraphics[width=0.12\linewidth]{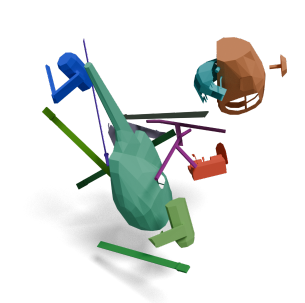}} &
        {\includegraphics[width=0.12\linewidth]{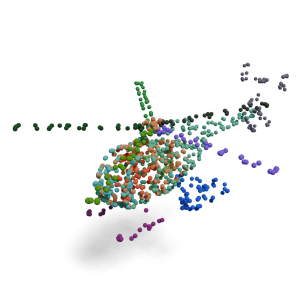}} &
        {\includegraphics[width=0.12\linewidth]{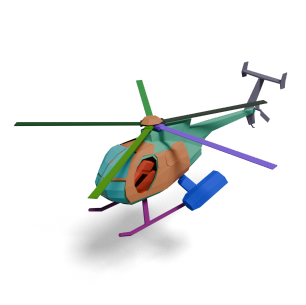}} &

        {\includegraphics[width=0.12\linewidth]{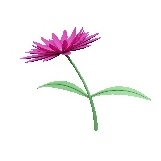}} &
        {\includegraphics[width=0.12\linewidth]{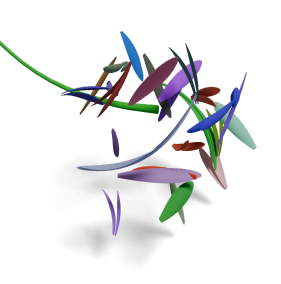}} &
        {\includegraphics[width=0.12\linewidth]{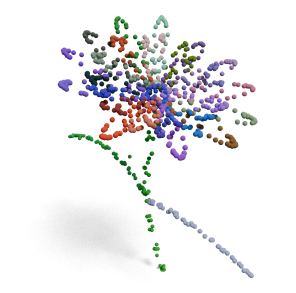}} &
        {\includegraphics[width=0.12\linewidth]{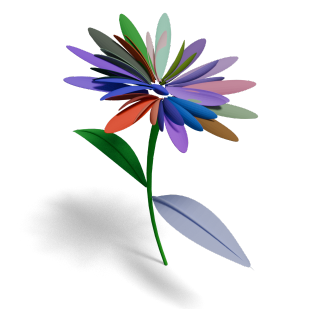}} \\

        {\includegraphics[width=0.12\linewidth]{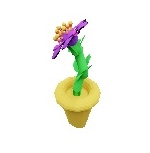}} &
        {\includegraphics[width=0.12\linewidth]{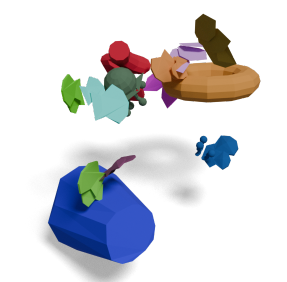}} &
        {\includegraphics[width=0.12\linewidth]{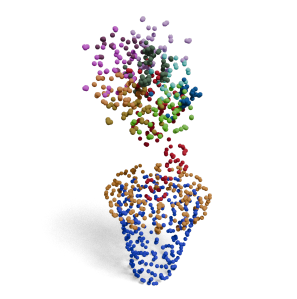}} &
        {\includegraphics[width=0.12\linewidth]{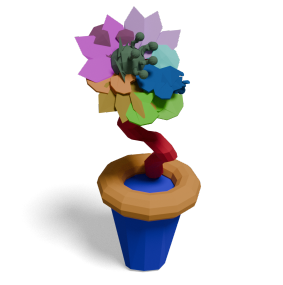}} &

        {\includegraphics[width=0.12\linewidth]{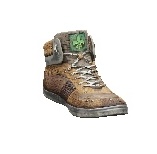}} &
        {\includegraphics[width=0.12\linewidth]{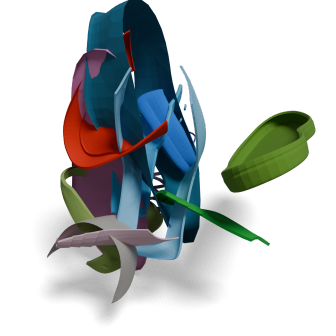}} &
        {\includegraphics[width=0.12\linewidth]{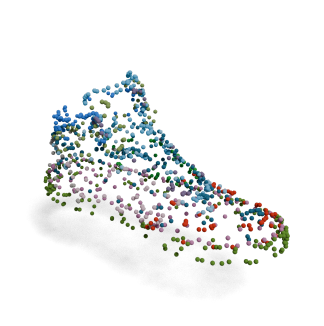}} &
        {\includegraphics[width=0.12\linewidth]{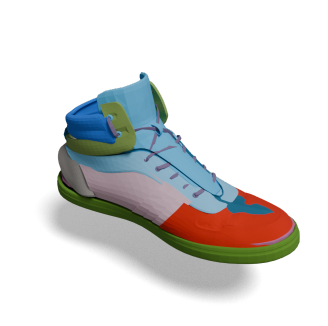}} \\

        {\includegraphics[width=0.12\linewidth]{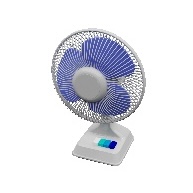}} &
        {\includegraphics[width=0.12\linewidth]{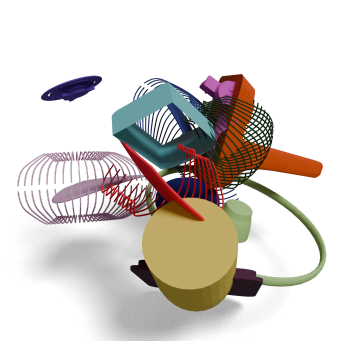}} &
        {\includegraphics[width=0.12\linewidth]{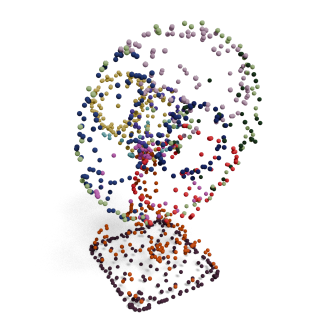}} &
        {\includegraphics[width=0.12\linewidth]{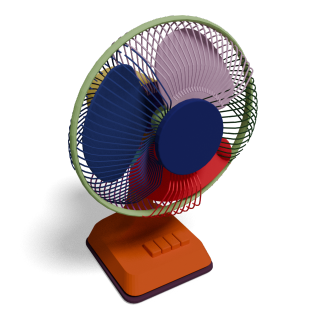}} &

        {\includegraphics[width=0.12\linewidth]{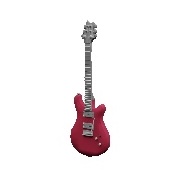}} &
        {\includegraphics[width=0.12\linewidth]{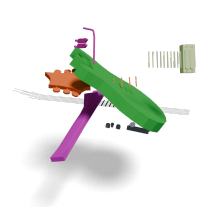}} &
        {\includegraphics[width=0.12\linewidth]{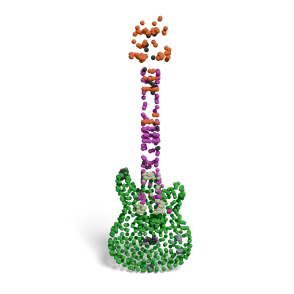}} &
        {\includegraphics[width=0.12\linewidth]{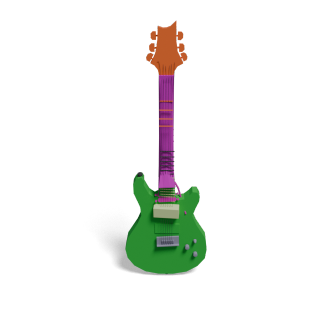}} \\

        Input Image & Input Parts & Anchor Points & Assembly & Input Image & Input Parts & Anchor Points & Assembly \\
    \end{tabular}
    \caption{More 3D part assembly results on Toys4K dataset.}
    \Description{More 3D part assembly results on Toys4K dataset.}
    \label{fig_only:toys4k}
\end{figure*}

To move beyond category-specific part assembly, we scale up our model in both capacity and data to enable general-purpose 3D part assembly in the wild. We train a DiT model with 195M parameters on our curated dataset comprising 320K diverse objects. Toys4K~\cite{stojanov2021using} is used as the test set and not included in the training data. Qualitative results are shown in Figure~\ref{fig:toys4k} and Figure~\ref{fig_only:toys4k}. Our trained model, for the first time, achieves automatic 3D part assembly for general objects. These objects span a wide range of daily categories including characters, creatures, plants, food, vehicles, etc. Moreover, they feature a large number of diverse 3D parts with complex geometries, semantics and their intricate relationships, offering a significant challenge for automatic assembly. Our model, thanks to its scalable designs on representation, model, and data, generates reasonable 3D assemblies in various scenarios, demonstrating great potentials in downstream applications such as 3D content creation, manufacturing, and robotics.

\subsection{Ablation Study}
\begin{table}[tb]
    \centering
    \caption{Ablation studies on PartNet Chair category.}
    \vspace{-5pt}
    \scalebox{0.9}{
    \begin{tabular}{lcccc}
        \toprule
        & SCD ↓ & PA ↑ & CA ↑ & SR ↑ \\
        \midrule
        Cross-attention & 19.44 & 24.82 & 9.09 & 0.00 \\
        w/o CFG & 8.54 & 64.12 & 51.43 & 23.99 \\
        w/o Coord \& Part ID & 7.82 & 63.84 & 51.50 & 22.16 \\
        256 Anchor Points & 7.95 & 63.23 & 51.24 & 19.77 \\
        Ours-img-PartNet & \textbf{6.08} & \textbf{71.01} & \textbf{57.55} & \textbf{30.08}\\
        \midrule
        Ours-img & \underline{7.21} & \underline{66.17} & \underline{52.20} & \underline{24.92} \\
        \bottomrule
    \end{tabular}}
    \label{tab:ablation}
    \vspace{-5pt}
\end{table}
We conduct ablation studies to validate the effectiveness of our design choices. All ablations are tested on the Chair category of the PartNet dataset, summarized in Table~\ref{tab:ablation}. \textit{Cross-attention} indicates replacing the concatenation of part features and noise tokens with cross attention to inject the part information, which cannot ensure the critical per-point alignment, thus fails to produce reasonable assembly. Additionally, reducing anchor points (256 v.s. 1024) leads to downgraded results, since it weakened the representation accuracy of both input and assembled shapes. To investigate the effect of scaling training data, we train a category-agnostic model (\textit{Ours-img-PartNet}) with all PartNet data, and test it on the Chair category. Although these additional data contain no chair, they provide common knowledge of objects and assembly, thus gives consistently better assemblies. Also, the classifier-free guidance (CFG) helps to emphasize the image information for disambiguating the assembly process. The concatenation of input part coordinates and their part indexes helps to complement the part features for better assembly.

\subsection{Part-aware 3D Generation Example}
\begin{figure*}[hb]
    \centering
    \includegraphics[width=1.0\linewidth]{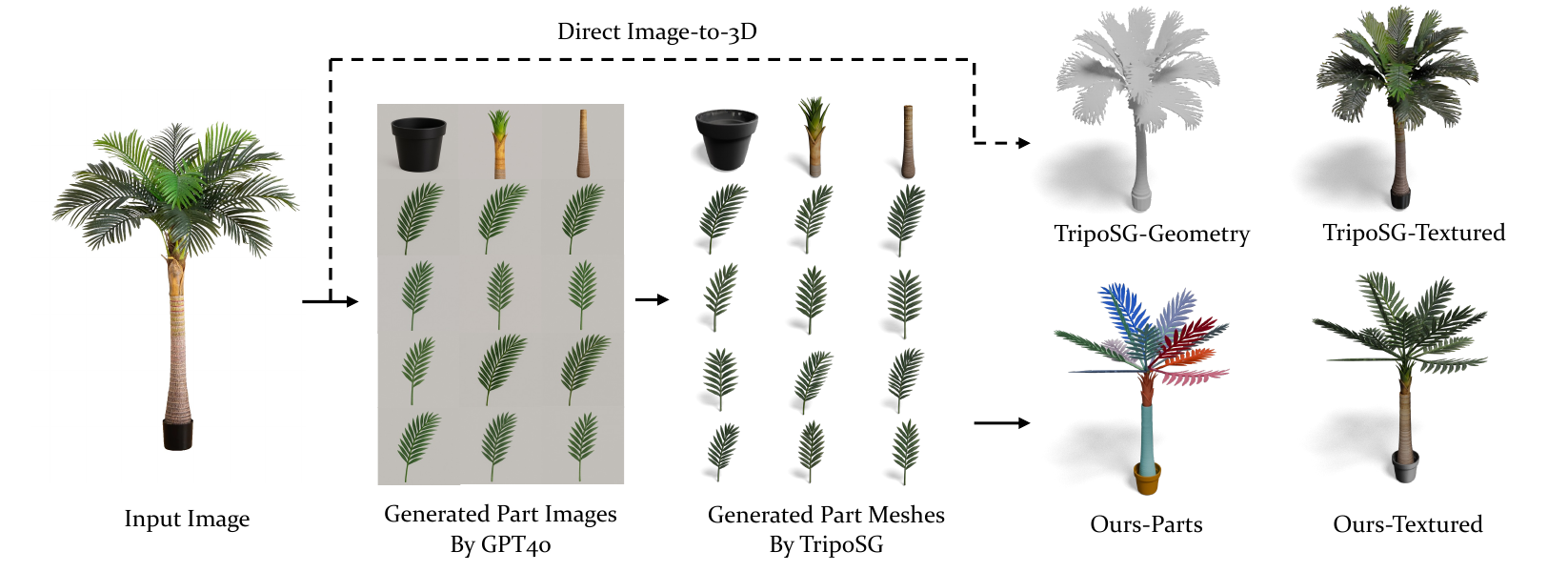}
    \caption{Example of our part-aware 3D generation prototype system.}
    \Description{Example of our part-aware 3D generation prototype system.}
    \label{fig:application}
\end{figure*}

With the proposed Assembler, we introduce a part-aware 3D generation prototype, as described in Sec.~\ref{sec:application}. Here we show an example of this pipeline in Figure~\ref{fig:application}. Starting from an input image, we first use GPT-4o~\cite{hurst2024gpt} to infer the parts and generate reference images for each part. These are then passed to an image-to-3D generator, such as TripoSG~\cite{li2025triposg}, to produce individual 3D part meshes. Our Assembler then takes in the 3D part meshes and the input image, to generate the complete 3D object model. As a comparison, we also employ the TripoSG to directly generate the 3D model from the input image. As shown in Figure~\ref{fig:application}, the directly generated 3D model struggles with fine-grained details (e.g., leaves) and lacks part-level structure. In contrast, our pipeline decomposes the task into high-resolution, part-specific generations (top-down), followed by part-aware assembly (bottom-up). This results in high-quality, modular 3D assets that support downstream editing and interaction. While the multi-stage pipeline may introduce cumulative errors and lower fidelity than direct generation, it demonstrates a new promising path towards user-friendly 3D content creation.

\subsection{Limitations and Future Works}
While Assembler marks a novel and valuable step toward general 3D part assembly, the problem remains far from solved. The model occasionally struggles in challenging scenarios involving numerous small parts or requiring precise boundary alignments such as LEGO-like structures. Future improvements may come from scaling up model capacity and training data, adopting more advanced generative techniques (e.g., rectified flows), and incorporating stronger priors (e.g., pre-trained point cloud diffusion models or object-centric graphs) to enhance robustness. Extending Assembler to 3D compositional scene generation, or enabling Assembler to handle missing or extraneous parts, are both interesting directions for future works.
\section{Conclusion}
In this paper, we present Assembler, a novel and scalable framework for general 3D part assembly. A shape-centric, anchor point-based assembly representation and its diffusion generative model are introduced to unlock the scalability, accompanied by a large-scale 3D part assembly dataset constructed via a simple yet effective data synthesis pipeline. Assembler is the first framework to demonstrate high-quality, automatic 3D part assembly for diverse, in-the-wild objects. Built on top of this foundation, we further introduce a part-aware 3D generation prototype that assembles modular, high-resolution 3D models from images, opening new possibilities for compositional 3D content creations.


\clearpage
\appendix
\setcounter{page}{1}

\section{More Details of Dataset Curation}
As described in the main paper, our data processing pipeline consists of four steps, namely filtering, segmentation, grouping and augmentation. The filtering step is mainly based on the captions, and we remove all the objects which contain any of ``scene, building, construction, area, village, ruined, house, rooms'' in captions. After that, we check the connected-components of mesh faces, and segment each connected component as a part. The mesh with dominant part (larger than 98\% of the total face areas) is also filtered out, since they are not beneficial for assembly training. Once getting the segmented parts, we first merge small parts into their neighbors, and employ KNN clustering to group the part units into larger, more semantic parts. To enrich the part diversity, we apply three different levels for the grouping, from the extensive grouping with less resulting part numbers (3-20 parts), to conservative grouping with most of the parts preserved (10-100 parts). Finally, each part is randomly rotated and translated to have an initial pose for assembly. 

\begin{figure*}[h]
    \centering
    \small
    \setlength{\tabcolsep}{0pt}
    \begin{tabular}{cccccccc}

        \centering
        {\includegraphics[width=0.12\linewidth]{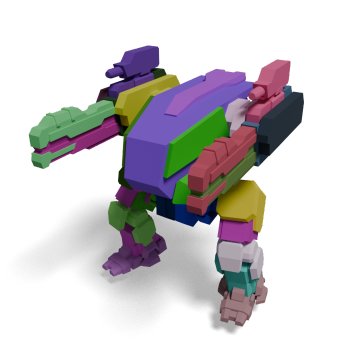}} &
        {\includegraphics[width=0.12\linewidth]{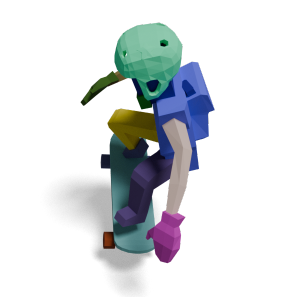}} &
        {\includegraphics[width=0.12\linewidth]{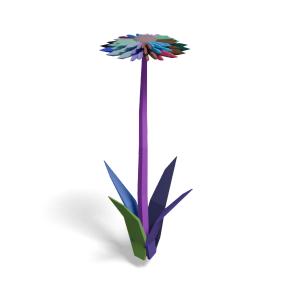}} &
        {\includegraphics[width=0.12\linewidth]{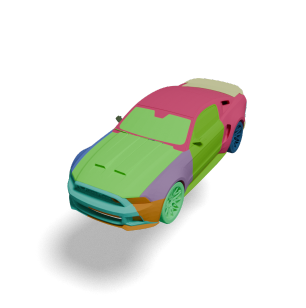}} &

        {\includegraphics[width=0.12\linewidth]{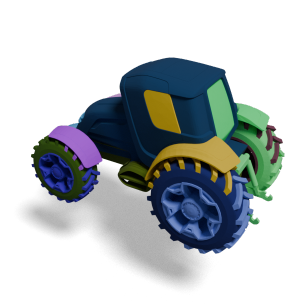}} &
        {\includegraphics[width=0.12\linewidth]{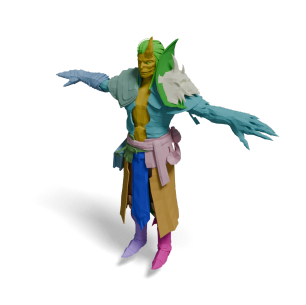}} &
        {\includegraphics[width=0.12\linewidth]{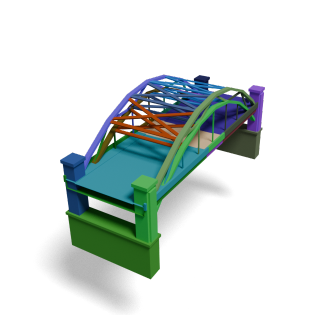}} &
        {\includegraphics[width=0.12\linewidth]{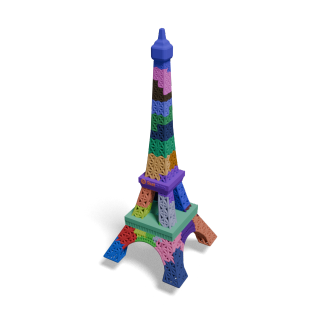}} \\

        {\includegraphics[width=0.12\linewidth]{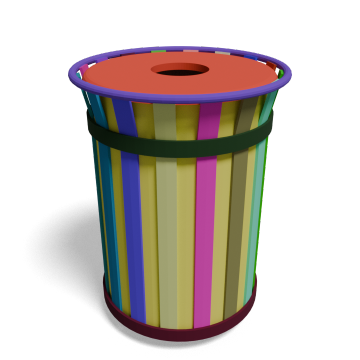}} &
        {\includegraphics[width=0.12\linewidth]{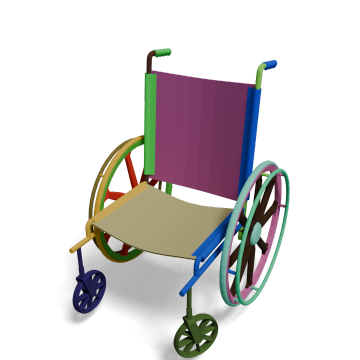}} &
        {\includegraphics[width=0.12\linewidth]{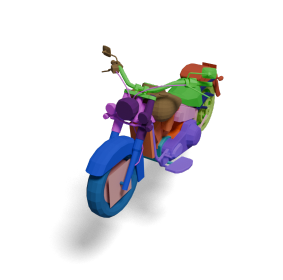}} &
        {\includegraphics[width=0.12\linewidth]{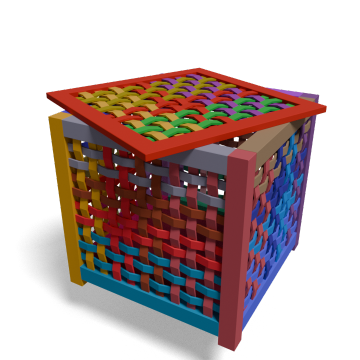}} &

        {\includegraphics[width=0.12\linewidth]{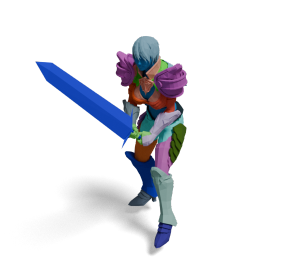}} &
        {\includegraphics[width=0.12\linewidth]{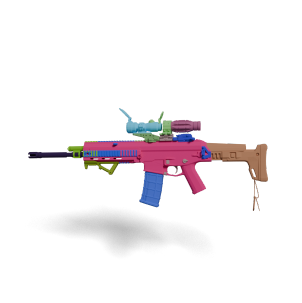}} &
        {\includegraphics[width=0.12\linewidth]{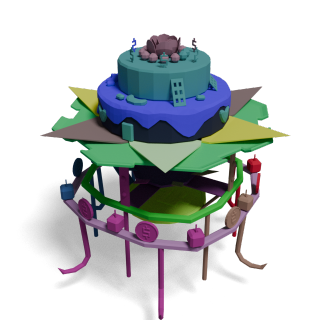}} &
        {\includegraphics[width=0.12\linewidth]{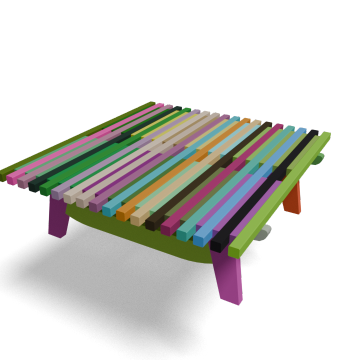}} \\

    \end{tabular}
    \caption{Examples from constructed 3D part assembly dataset.}
    \label{fig:supp_dataset}
\end{figure*}

Using the above data processing pipeline, we process the data from PartNet, TRELLIS-500K~\cite{xiang2024structured} collection, namely Objaverse Sketchfab~\cite{deitke2023objaverse}, Objaverse Github~\cite{deitke2023objaverse}, ABO~\cite{collins2022abo}, HSSD~\cite{khanna2024habitat}, 3D-FUTURE~\cite{fu20213d}, and ShapeNet~\cite{chang2015shapenet}. Table~\ref{tab:dataset} lists the detailed object numbers before and after processing. In total, we curate over 320K high-quality objects with their part assemblies. Different levels of grouping and the random part pose augmentation further lift the scale of part assemblies for training. In Figure~\ref{fig:supp_dataset}, we show some examples from the dataset, containing diverse objects with clean part segmentations.
\section{More Implementation Details}
We train the anchor point DiT model with two configurations. One is DiT-S with 16 layers, 384 hidden size, and 8 heads, resulting in 49M parameters, which is trained for category-specific comparison on PartNet. Another is DiT-M with 16 layers, 768 hidden size and 8 heads, resulting in 195M parameters, used for training on our curated dataset. For training, we adopt a corase-to-fine training strategy, where we first train with 256 token length (anchor point numbers) for 500K steps, followed by another 200K steps with 1024 tokens. This accelerate the overall diffusion training. Our DiT-M model is trained with 8 Nvidia A100 GPUs for around five days. As for the condition, we employ the DINOv2-ViT-B14 as the image encoder, and Dora VAE 1.1 to extract shape features for sparse anchor points. For the assignment of anchor points on each part, we first calculate the size ratios of each part of a object with their face areas, then proportionally assign anchor point quota to each part under the guarantee of a minimum 10 points for a part. For each part, the sampled dense point cloud and the anchor points are input to the Dora VAE encoder as values and queries, to extract shape features for the anchor points. Our code will be released.

\section{More Details of Category-specific Comparison}
We compare with representative open-source 3D part assembly methods DGL~\cite{zhan2020generative}, RGL~\cite{narayan2022rgl}, Score-PA~\cite{cheng2023score} and SPAFormer~\cite{xu2025spaformer}. We use their released code, data and checkpoints to evaluate. Official test split of each category is used for evaluation. Our Assembler is evaluated on the processed data of SPAFormer, to ensure the fairness. All the input parts are canonicalized using PCA analysis, following these baselines. DGL and RGL only have valid checkpoints for Chair and Lamp categories, so we test them on these two categories. Other related methods (3DHPA~\cite{du2024generative}, CCS~\cite{zhang2024scalable}, IET~\cite{zhang20223d}) have not released their codes or models, thus we cannot directly compare with them.

\begin{table}[t]
    \centering
    \caption{Statistics of our curated 3D part assembly dataset.}
    \scalebox{0.9}{
    \begin{tabular}{lrrr}
        \toprule
        Source & \# Original Objects & \# Objects After Filtering \\
        \midrule
        PartNet & 16,995 & 16,995\\
        Objaverse Sketchfab & 168,307 & 91,786\\
        Objaverse Github & 311,843 & 156,526\\
        ABO & 4,485 & 3,926\\
        HSSD & 6,670 & 5,611\\
        3D-FUTURE & 9,472 & 8,586\\
        ShapeNet & 54,217 & 37,226\\
        \midrule
        Total &  571,989 & 320,656 & \\
        \bottomrule
    \end{tabular}}
    \label{tab:dataset}
\end{table}
\section{More Details of Part-aware 3D Generation Pipeline}
Our part-aware 3D generation pipeline consists of three modules, VLMs for inferring and generating part images, image-to-3D generator for creating part meshes, and our Assembler for part assembly. We employ the GPT-4o~\cite{hurst2024gpt} due to its unified multi-modal understanding and generation capability. For image-to-3D generation, we employ the state-of-the-art open-source model TripoSG~\cite{li2025triposg}. Other alternatives could also be used. For GPT-4o, we send the original input image to it, and prompt it with \textit{"Assume yourself as a 3D artist. Given a reference image, you need to create the corresponding 3D object. For the object in this picture, please first reason and separate all the parts of the object. Each part should be the smallest unit with semantics. Based on the original picture, generate an image for each part. Try to retain or enhance the details in the original picture as much as possible."} Each generated part image is in 1024x1024 resolution with great details. We keep the GPT-4o to generate the next part until it marks all the parts are generated. Then, all the part images are input into TripoSG to generate part meshes one by one. With all these part meshes and the original input image, our Assembler then automatically generate the complete object. Since TripoSG can also generate textures on each part, and the Assembler will retain all these textures of each part mesh, we can produce a textured, high-resolution, part-aware object mesh.

\clearpage
\bibliographystyle{ACM-Reference-Format}
\bibliography{bib}

\end{document}